\documentclass{article}
\usepackage[utf8]{inputenc}
\usepackage{amsmath,amssymb}
\usepackage{graphicx}
\usepackage{algorithm}
\usepackage{algorithmic}
\usepackage{tikz}
\usepackage{authblk}

\DeclareMathOperator{\EX}{\mathbb{E}}

\title{Providing theoretical learning guarantees to Deep Learning Networks}
\author[1]{Rodrigo F. de Mello}
\author[1]{Martha Dais Ferreira}
\author[1]{Moacir A. Ponti}

\affil[1]{University of S\~{a}o Paulo, Institute of Mathematics and Computer Sciences, Department of Computer Sciences, S\~{a}o Carlos, SP, Brazil\\\{mello,daismf,moacir\}@icmc.usp.br}

\date{November 28th, 2017}

\begin{document}

\maketitle

\begin{abstract}
Deep Learning (DL) is one of the most common subjects when Machine Learning and Data Science approaches are considered. There are clearly two movements related to DL:
the first aggregates researchers in quest to outperform other algorithms from literature, trying to win contests by considering often small decreases in the empirical risk; and the second investigates overfitting evidences, questioning the learning capabilities of DL classifiers.
Motivated by such opposed points of view, this paper employs the Statistical Learning Theory (SLT) to study the convergence of Deep Neural Networks, with particular interest in Convolutional Neural Networks. In order to draw theoretical conclusions, we propose an approach to estimate the Shattering coefficient of those classification algorithms, providing a lower bound for the complexity of their space of admissible functions, a.k.a. algorithm bias. Based on such estimator, we generalize the complexity of network biases, and, next, we study AlexNet and VGG16 architectures in the point of view of their Shattering coefficients, and number of training examples required to provide theoretical learning guarantees. From our theoretical formulation, we show the conditions which Deep Neural Networks learn as well as point out another issue: DL benchmarks may be strictly driven by empirical risks, disregarding the complexity of algorithms biases.
\end{abstract}

\section{Introduction}

Classification algorithms infer decision boundaries on some input space, so examples under different labels are correctly shattered into different regions.
In this context, \emph{learning} means inferring a classifier $f:X \rightarrow Y$ from training pairs $(x_i, y_i) \in X \times Y$, in which:
\begin{enumerate}
        \item $X = \{x_1, \ldots, x_n\}$ is the input space;
        \item $Y = \{y_1, \ldots, y_n\}$ is the output space.
\end{enumerate}

Every $x_i \in X$ represents some object in terms of its attributes, which must be representative enough to allow the proper learning of a concept.
Each $y_i \in Y$ refers to a label, class or category to be learned given examples in $X$.
The estimated classifier $f$ must operate on unseen examples during the training stage, and yet provide correct labels as output.
This is commonly referred to as generalization in the context of Machine Learning (ML)~\cite{vonLuxburg}.

In ML, we wish to find an algorithm that, provided enough data, outputs a classification hypothesis $f$ with the small as possible error computed according to some loss function $\ell(f(x_i), y_i)\, \forall i$~\cite{Vapnik2013nature}.
Cross Entropy and Squared-Error are the most commonly used loss functions in the Deep Learning scenario~\cite{Ponti2017everything,Goodfellow2016deep}. The loss function is fundamental to guide convergence to the best as possible $f$ contained in some algorithm bias $\mathcal{F}$, however that is not sufficient to ensure such selected function provides good enough results over unseen data. This is the main point addressed by the Statistical Learning Theory (SLT) through the Empirical Risk Minimization Principle (ERMP).

Two main parts are fundamental for the study of learning: i) the first is ERMP, which approximates the true (and unknown) loss function by taking only the observed examples (from the training set), and uses the hypothesis that minimizes the error inside the training set; ii) the second involves finding a trade-off between the complexity of the hypothesis space, i.e., the bias of the classification algorithm, and the classification error computed on the training data (the empirical risk).

According to SLT, the following are necessary to ensure learning~\cite{Vapnik2013nature}: i) no assumption is made about the joint probability function $P(X \times Y)$, so that no prior knowledge about the association between input and output variables is required; ii) examples must be sampled in an independent manner since SLT relies on the Law of Large Numbers; iii) labels may assume non-deterministic values due to noise and class overlapping, and learning may still happen; iv) $P(X \times Y)$ is fixed, i.e., static, so it does not change along time, what is also required by the Law of Large Numbers; v) $P(X \times Y)$ is unknown at the training stage, therefore estimated using collected samples.

By relying on such assumptions and on the Law of Large Numbers, Vapnik~\cite{Vapnik2013nature} proved the following bound for \emph{supervised learning algorithms}:
\begin{align}\label{eq:chernoff}
\begin{split}
   P(\sup_{f \in \mathcal{F}}|R(f) & - R_\text{emp} (f)| > \epsilon) \leq \\
   & 2 P(\sup_{f \in \mathcal{F}}|R'_\text{emp}(f) - R_\text{emp} (f)| > \epsilon/2) \leq  \\
   & \qquad \qquad \qquad 2 \mathcal{N}(\mathcal{F},2n) \exp( -n \epsilon^2/4 ),
\end{split}
\end{align}
in which $\mathcal{F}$ is the space of admissible functions for some classification algorithm also known as bias, $R(f)$ is the expected risk of some classifier $f$ provided the joint distribution $P(X \times Y)$ is known as defined in Equation~\ref{eq:rexp}, $R_\text{emp}(f)$ is the empirical risk measured on a given sample drawn from the joint distribution as seen in Equation~\ref{eq:remp}, $\epsilon$ is an acceptable divergence between those risks, $R'_\text{emp}(f)$ is also an empirical risk but computed on second sample from the same joint distribution, and $\mathcal{N}(\mathcal{F},2n)$ is the cardinality of $\mathcal{F}$ also known as \emph{Shattering coefficient}, i.e., the number of distinct functions contained in the algorithm bias, provided $2n$ examples. The right-most term of such inequality comes from the Chernoff bound~\cite{vonLuxburg, Devroye96}.

\begin{equation}\label{eq:rexp}
    R(f) = \EX(\ell(f(X), Y))
\end{equation}

\begin{equation}\label{eq:remp}
    R_\text{emp}(f) = \frac{1}{n} \sum_{i=1}^{n} \ell(f(x_i), y_i)
\end{equation}

As result of such proof, Vapnik~\cite{Vapnik2013nature} obtained the Generalization bound:
\begin{equation}\label{eq:generalizationbound}
 R(f_{\text{w}}) \leq \underbrace{R_\text{emp} (f_{\text{w}})}_\text{training error} + \underbrace{\sqrt{-\frac{4}{n} \left( \log(\delta) - \log(2\mathcal{N}(\mathcal{F},2n)) \right)}}_{\text{divergence factor}\; \epsilon},
\end{equation}
in which $f_{\text{w}}$ is the worst possible classifier contained in $\mathcal{F}$. Notice the training error is represented in terms of the empirical risk and the additional term is associated with the number of admissible functions in space $\mathcal{F}$, defining an upper bound for the actual risk, i.e., for unseen examples. As a lemma, learning occurs only if:
\begin{equation}
  \lim_{n \rightarrow \infty} \frac{\log \mathcal{N}(\mathcal{F},2n)}{n} = 0,
\end{equation}
therefore, the Shattering coefficient $\mathcal{N}(\mathcal{F},2n)$ is mandatory to prove under which conditions any classification algorithm learns according to the SLT.

This paper addresses the analysis of Deep Learning (DL) methods for classification tasks, in particular image classification. Deep Neural Networks (DNNs) are state-of-the-art in many benchmark datasets and were also part of solution of important visual recognition challenges such as the ILSVRC (ImageNet)~\cite{ILSVRC15}. Due to that, the architectures and models trained on those datasets are currently used as off-the-shelf methods in many applications.
On the other hand, recent studies showed that DNNs can easily fit both random labels and random input data via a type of brute-force memorization, raising questions about what those models are actually learning.

Zhang et al.~\cite{Zhang2016understanding} showed that deep networks can reach zero training error (empirical risk) in CIFAR-10 and ImageNet-1000 benchmark datasets. This zero empirical risk is indeed common in experiments reported in the Deep Learning literature and, alone, is an indicative of overtraining. In addition to that, the authors used the same training data however randomizing the labels so that to break the correspondence between the labels and the actual contents of the image, i.e., randomizing the joint distribution $P(X \times Y)$. After training with random labels, the networks also produced zero training error. A similar behaviour is found when training images with noise~\cite{Zhang2016understanding}. Nazare et al.~\cite{Nazare2017deep} showed Convolutional Neural Networks (CNNs) do not generalize well for different types and levels of noise in images, in fact those are able to fit even visually unrecognizable noisy images, learning the specific type and level of noise so that in test stage the clean images or images with other types of noise are misclassified. The relationship between the network capacity and the input noise is also studied in~\cite{Arpit2017closer}, indicating there is dependency between the data and the network capacity needed for such task.

The aforementioned studies motivate the study of generalization and convergence for Deep Networks. Based on the theoretical framework provided by SLT, we claim Deep Learning Networks should be studied and their convergence analyzed in order to prove under which conditions learning is ensured. During the last years, researchers and companies have been focusing on such approaches to tackle a great variety of classification tasks, being, sometimes, even biased towards adopting them without questioning their pros and cons~\cite{krizhevsky2012imagenet, szegedy2015going, zeiler2014visualizing, simonyan2014very, he2016deep}. Motivated by such scenario, this paper introduces an approach to estimate the Shattering coefficient of classification algorithms, providing a lower bound for the complexity of their space of admissible functions, a.k.a. algorithm bias.

Based on such estimator, we generalize the bias complexity of Deep Learning Networks, and yet study AlexNet and VGG16 architectures in the point of view of their Shattering coefficients. From our theoretical formulation, we show the conditions which Deep Neural Networks learn as well as point out another issue: DL competitions may be strictly driven by empirical risks, disregarding the complexity of algorithms biases.

As main contribution, we propose an algorithm to estimate the Shattering coefficient of Convolutional Neural Networks, and employ it to analyze the complexity of CNN biases. In addition, we study the learning guarantees of CNNs in practical scenarios. All conclusions rely on the Statistical Learning Theory (SLT) and confirm the importance of investigating the divergence between the empirical and the expected risks in terms of the Generalization bound.

This paper is organized as follows: Section~\ref{sec:shattering} considers the Shattering coefficient to analyze the complexity of algorithm biases, as well as estimate such function in practical scenarios; Convolutional Neural Networks are formalized in terms of their processing units and how they operate on the data space in Section~\ref{sec:cnn}; we employ such formalization in Section~\ref{sec:general} in order to propose a general methodology to quantify the Shattering coefficients for Convolutional Neural Netwoks; AlexNet and VGG16 architectures are studied using the MNIST and the ImageNet datasets in terms of their Shattering coefficients and, consequently, their convergences and learning guarantees in Section~\ref{sec:analyzing}; Section~\ref{sec:discussion} discusses about the effects of different Shattering coefficients in the light of the related work; Finally, conclusions are presented in Section~\ref{sec:conclusions}.

\section{Estimating the Shattering coefficient}\label{sec:shattering}

In order to proceed with the convergence analysis of Deep Neural Networks, we propose an estimator for the Shattering coefficient of a single neuron as listed in Algorithm~\ref{alg:shattering-estimator}. It builds up a single $(\mathbb{R}-1)$-hyperplane on an $\mathbb{R}$-dimensional space and counts up how many different classifications (using hashtable \emph{shatterWays}) are obtained for a Normally distributed set of examples (command \emph{rnorm} requires the average, the standard deviation and the sample size $n$). The finite number of iterations, parameter \emph{iter}, is used to compose a power of $n$, making the algorithm run more times as the sample size increases, which is mandatory to allow more distinct classifications. As matter of fact, given such examples are organized in the input space using a single probability distribution, and due to the finite number of iterations set by the user, this will always be an underestimator for the Shattering coefficient, meaning it works as a lower bound for the total number of admissible functions in $\mathcal{F}$. Therefore, if even for such lower bound the classification algorithm requires many data examples to converge, so it will be even worse for more complex dataset organizations.

Still detailing Algorithm~\ref{alg:shattering-estimator}, observe parameters \emph{start} and \emph{end} set the initial and final sample sizes to be considered, \emph{minValue} and \emph{maxValue} define the bounds for the Uniform probability distribution used to randomly initialize possible hyperplanes as $x \mathbf{w} + b$. Every random hyperplane is then used to shatter the space and produce the corresponding labels $\{+1,-1\}$. If such the hyperplane applied to some example outputs a value greater than or equal to zero, than the resulting label is considered positive, otherwise negative. Vector \emph{labels} contains $+1$s and $-1$s which are then concatenated to form a hash key used to indicate when a given classification was performed (command \emph{shatterWays.put(key, TRUE)}). Pairs \emph{(n,shatterWays.size())} stored in Matrix \emph{estimation} contain the sample size followed by the number of distinct classifications found. A polynomial function is fitted on Matrix \emph{estimation} to represent the relation between sample size and distinct classifications.

\begin{algorithm}[h!]
\caption{Shattering estimator for a single neuron.}\label{alg:shattering-estimator}
\begin{algorithmic}

    \REQUIRE iter $> 0$, start $\geq 1$, end $>$ start, R $\geq 1$, average, stdev, minValue, maxValue
	\STATE estimation = Matrix()
	\FOR{n in start:end}
		\STATE x = Matrix()
		\FOR {j in 1:R}
			\STATE x = [ x, Normal(average, stdev, n) ]
		\ENDFOR

		\STATE shatterWays = Hashtable()

		\FOR {j in 1:pow(n, iter)}
			\STATE ${\bf w}$ = Uniform(minValue, maxValue, R)
			\STATE b = Uniform(minValue, maxValue, 1)

			\STATE labels = x ${\bf w} +$ b
			\STATE positiveIds = which(labels $>= 0$)
			\STATE negativeIds = which(labels $< 0$)
			\STATE labels[positiveIds] = $1$
			\STATE labels[negativeIds] = $-1$

			\STATE key = concat(labels)
			\STATE shatterWays.put(key, TRUE)
		\ENDFOR
		
		\STATE estimation[n,] = [ n, shatterWays.size() ]
		
	\ENDFOR

	\STATE return shatter
    
\end{algorithmic}
\end{algorithm}

\begin{figure}
	\centering
	\includegraphics[width=0.85\linewidth]{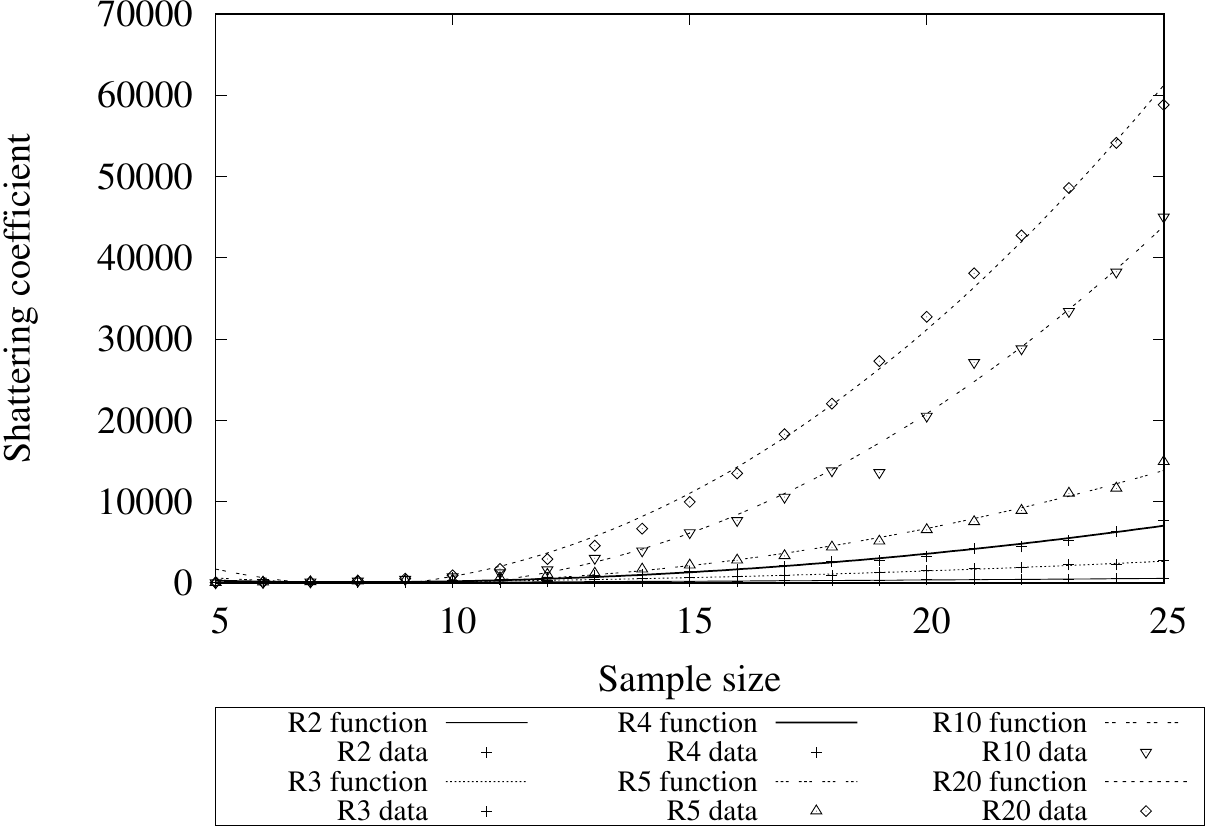} 
	\caption{Shattering coefficients for different algorithm biases.}
	\label{fig:image1}
\end{figure}

Using Algorithm~\ref{alg:shattering-estimator}, we estimated the Shattering coefficients for multidimensional input spaces as listed in Table~\ref{tab:shatterings}, setting \emph{iter}$=1,000$, \emph{start}$=1$, and \emph{end}$=100$. Notice the significant changes on the polynomial coefficients, which are related to the complexity of algorithm biases. Figure~\ref{fig:image1} shows those curves in terms of distinct classifications versus sample size $n$. Observe all curves visually present a good fitting, as confirmed by the squared of residuals and error percentage.

\begin{table}[h!]
	\centering
	\caption{Shattering coefficients of a single neuron while classifying input spaces under a different number of dimensions.}
	\begin{tabular}{c|c|c}
		\hline
		Dimensions  & Shattering coefficients  & Squared of residuals\\
		&                          & (Error percentage)\\ 
		\hline
		$\mathbb{R}^2$  & $0.91 n^2-0.98 n + 2.68$          & $6.94 (5.07)$    \\
		$\mathbb{R}^3$  & $7.22 n^2 -87.01 n + 311.97$      & $66.38 (6.13)$   \\
		$\mathbb{R}^4$  & $23.51 n^2 -366.96 n + 1509.07$   & $236.83 (6.73)$  \\
		$\mathbb{R}^5$  & $50.97 n^2 -868.62 n + 3671.67$   & $420.86 (5.51)$  \\
		$\mathbb{R}^{10}$ & $166.80 n^2 -2895.02 n + 11995.6$ & $1249.48 (5.00)$ \\
		$\mathbb{R}^{20}$ & $198.51 n^2 -2919.44 n + 10128.6$ & $1080.02 (3.63)$ \\
		\hline
	\end{tabular}
	\label{tab:shatterings}
\end{table}

Now we discuss some scenarios and employ the Shattering coefficients to understand convergence. 
Firstly, consider a hidden layer of a Multilayer Perceptron with $2$ neurons and let the input space be $\mathbb{R}^2$.
Due to the presence of $2$ neurons, two linear hyperplanes will be found after the training stage.
Each one with the following Shattering coefficient: $f(n)=0.91 n^2-0.98 n + 2.68$.
By having two neurons, the resulting Shattering of such layer is $f(n)^2$, given they can be reorganized such as in a cross shape and double the shattering possibilities.

\begin{figure}
	\centering
	\includegraphics[width=1\linewidth]{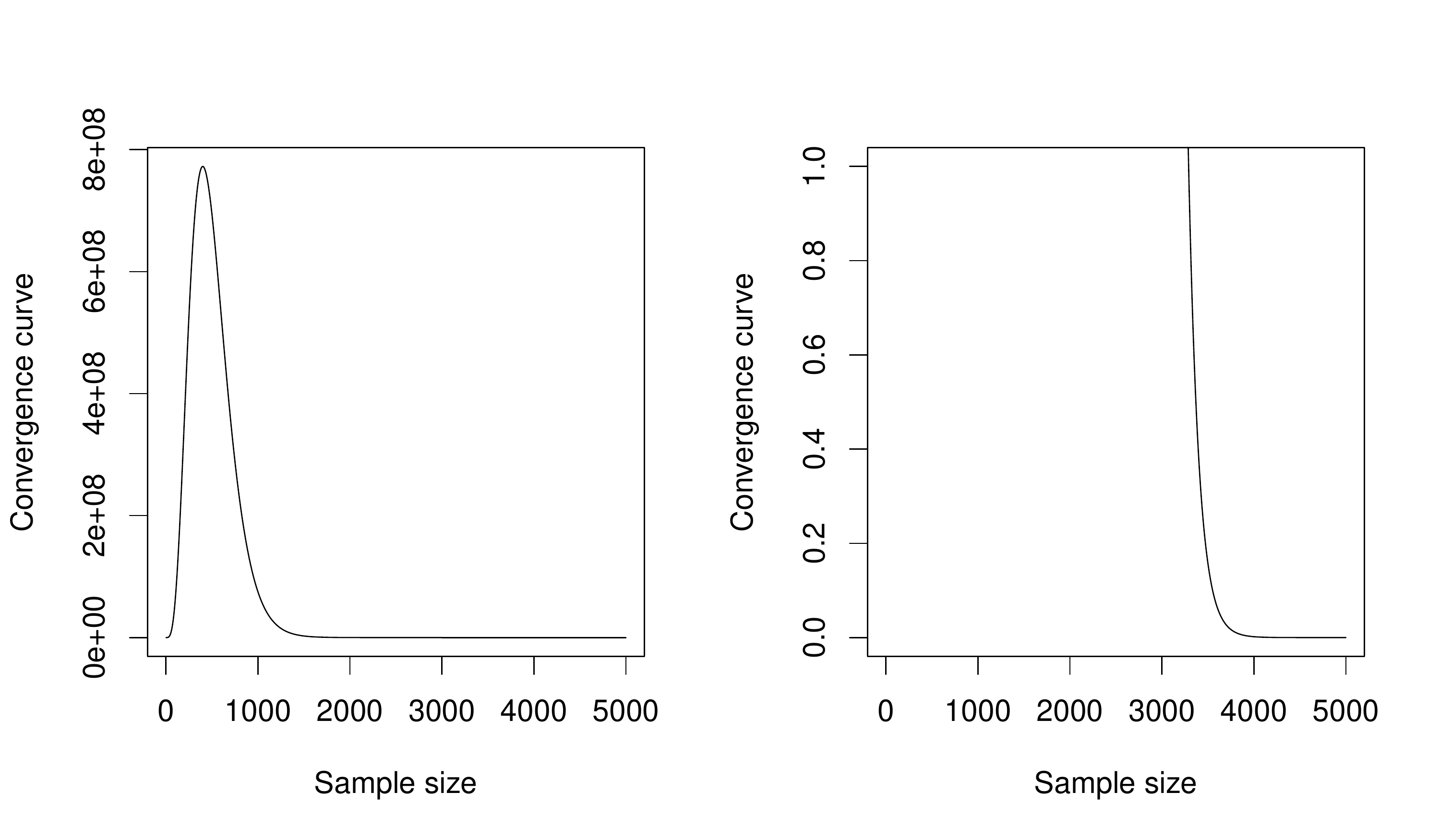}   
	\caption{The convergence curve according to the probability value below $0.05$ requires a sample size $n=3,629$.}
	\label{fig:image2}
\end{figure}

From the Statistical Learning Theory, the algorithm convergence is given by:
\begin{align*}
\begin{split}
 2 \mathcal{N}(\mathcal{F},2n) & \exp( -n \epsilon^2 ) = \\
 & 2 f(n)^2 \exp( -n \epsilon^2 ) = \\
 &\qquad 2 (0.91 n^2-0.98 n + 2.68)^2 \exp( -n 0.1^2 ), 
\end{split}
\end{align*}
considering $\epsilon=0.1$, which measures the divergence between expected risk $R(f)$ and empirical risk $R_\text{emp}(f)$. The convergence to some probability value below $0.05$ (that is $5\%$ as typically considered by most Statistical hypothesis tests) requires a sample size $n=3,629$, which is presented by the obtained convergence curve plotted in Figure~\ref{fig:image2}.
Observe that as $\epsilon$ is reduced, a greater number of training examples will be required to ensure learning.

In summary, observe that by increasing the number of neurons $k$ at the hidden layer, the more training examples are necessary to ensure learning convergence.
Taking that into account, the Chernoff bound becomes:
\begin{align*}
 2 f(n)^k \exp( -n \epsilon^2 ) = \delta,
\end{align*}
assuming some acceptable $\delta$ as the probability for divergence between risks $R(f)$ and $R_\text{emp}(f)$, which is here set as $\delta=0.05$ as commonly used in Statistics.
To recall, as defined by Vapnik~\cite{Vapnik2013nature} learning only occurs when:
\begin{align*}
  \lim_{n \rightarrow \infty} \frac{\log \mathcal{N}(\mathcal{F},n)}{n} = 0,
\end{align*}
from which we can compute the sufficient sample size to ensure learning in form:
\begin{align*}
  \lim_{n \rightarrow \infty} \frac{\log f(n)^k}{n} = 0.
\end{align*}

For instance, in case of $f(n)=0.91 n^2-0.98 n + 2.68$, we have:
\begin{align*}
  \lim_{n \rightarrow \infty} \frac{\log f(n)^k}{n},
\end{align*}
as the general case for $k$-neurons at the hidden layer.
Therefore, the greater $k$ is, the more examples are required to ensure learning.

Provided $k=2$:
\begin{align*}
  \lim_{n \rightarrow \infty} \frac{\log (0.91 n^2-0.98 n + 2.68)^2}{n},
\end{align*}
as a consequence we can set some acceptable numerical threshold such as:
\begin{align*}
  \frac{\log (0.91 n^2-0.98 n + 2.68)^2}{n} \leq \gamma,
\end{align*}
thus after reaching $\gamma$, we have the minimum sample size $n$ to ensure convergence.

Table~\ref{tab:sample_number_hyperplanes} illustrates the sample size required to ensure $\gamma = 0.01$ for different number of neurons $k$.
This value is taken due to $\gamma$ is close enough to zero as expected from some convergent process.
All data from such a table were fitted in order to find a polynomial to characterize the sample size required to ensure learning provided the number of hyperplanes as follows:
\begin{align*}
g(h) = 9.16 h^2 + 1966.38 h -812.24, 
\end{align*}
so, provided the number of hyperplanes $h$, it estimates the necessary number of training examples. Notice the influence of the number of neurons at the hidden layer, even for a very simple input space $\mathbb{R}^2$, in which an approximation produced $212.37$ as squared of residuals with error percentage equals to $11.89\%$.

\begin{table}[h!]
\centering
\caption{Minimal training sample size to ensure learning convergence assuming $\gamma = 0.01$ and an input space $\mathbb{R}^2$.}\label{tab:sample_number_hyperplanes}
\begin{tabular}{c|c}
  \hline
  \# hyperplanes  & Sample size \\ 
  \hline
  1 & 1446\\
  2 & 3211\\
  3 & 5094\\
  4 & 7051\\
  5 & 9065\\
  10 & 19681\\
  20 & 42435\\
  30 & 66332\\
  \hline
\end{tabular}
\end{table}

\section{Convolutional Neural Networks}\label{sec:cnn}

Convolutional Neural Networks (CNNs or ConvNets) are probably the most well known Deep Learning model used to solve Computer Vision tasks, in particular image classification.
The most basic building blocks of CNNs are convolutions, pooling (downsampling) operators, activation functions, and fully-connected layers, which are essentially similar to hidden layers of a Multilayer Perceptron (MLP). Architectures such as AlexNet~\cite{krizhevsky2012imagenet}, VGGNet~\cite{simonyan2014very}, ResNet~\cite{he2016deep} and Inception~\cite{szegedy2015going} have become very popular.

A CNN is composed of a set of functions $f_l(.)$ (related to some layer $l$), each of them taking some input $I_l$ and a set of parameters $\Theta_l$ to output a transformed version of the input $I_{l+1}$, as follows~\cite{Ponti2017everything}:
\begin{align*}
    f(I) = f_L \left( \cdots f_2(f_1(I_1, \Theta_1);\Theta_2) \cdots), \Theta_L \right),
\end{align*}
in which functions $f_l(.)$ are the basic building blocks, most commonly convolutional layers. Such layers are composed of units (neurons) that perform convolution on the input data.

Convolutions can be seen as filters (also known as kernel matrices), each of those is represented by a $k \times k$-matrix of weights $\theta_i$. At first, those weights are randomly initialized, and during optimization process those will converge to values defining the range of frequencies of interest for each layer and neuron~\cite{gonzalez2007modeling}.
Those filters in the convolutional layer produces an affine transformation of the input, what turns out to be the same as a linear combination of neighborhood pixels defined by the filter size.
Each region that the filter processes is called local receptive field: an output value (pixel) is a combination of the input pixels in this local receptive field.
That makes the convolutional layer different from layers of a Multilayer Perceptron (MLP) network.
For example, in MLP, each neuron produces a single output based on all values from the previous layer, whereas in a convolutional layer, an output value $I(k,i,j)$ is based on a filter $k$ and local data coming from the previous layer centered at a position $(i,j)$.
Despite such a difference, a convolutional layer in fact produces a linear transformation of each local receptive field.

Each convolutional neuron is responsible for applying a convolution operation on input images. In fact, the correlation operation is mostly usually applied:
\begin{align*}
o_{i,j} = (I \otimes \theta)_{i,j} = \sum_{x=1}^{m} \sum_{y=1}^{n} I_{i+x,j+y} \theta_{x,y},
\end{align*}
in which $I$ is the input image, $\theta$ is the filter, $i$ and $j$ are indices of $I$, $x$ and $y$ are indices of $\theta$, and, finally, $o$ is the output image produced. Observe that this operation is an inner-product between the filter and every local receptive field, when both terms are represented as vectors. Such inner-product is summarized as follows:
\begin{align*}
o_{i,j} = \left \langle \vec{I}_{i,j}, \vec{\theta} \right \rangle,
\end{align*}
in which $\vec{I}_{i,j}$ represents the column vector considering the local receptive field of a pixel $i$ and $j$ of $I$, $\vec{\theta}$ is the column vector obtained from the filter, and $o_{i,j}$ corresponds to a pixel of the output image.

In this context, each neuron corresponds to an $m \times n$-linear hyperplane responsible for linearly shattering the data space formed from local receptive fields of input images~\cite{Scholkopf2002learning}. Each convolutional layer is composed of a set of neurons with different filter weights, therefore producing a variety of linear divisions on the input data space in attempt to support classification. The combination of half spaces produced after such hyperplanes allows to represent nonlinear functions.

It is also worth to study convolutional neurons at a given layer as a linear transformation. Consider the filter of each neuron $k$ organized as a column vector $\vec{\theta_k}$, then a transformation matrix $\Theta$ is obtained after the concatenation of those column vectors:
\begin{align*}
\Theta^T\vec{I}_{i,j} = \left[
\begin{array}{l}
\vec{\theta_1}\\
\vdots\\
\vec{\theta_l}
\end{array}
\right]^T
\vec{I}_{i,j} =
\left[
\begin{array}{l}
o_{i,j}^1\\
\vdots\\
o_{i,j}^l
\end{array}
\right] = \vec{o}_{i,j},
\end{align*}
in which $l$ is the number of neurons, $\vec{I}_{i,j}$ corresponds to a column vector positioned at the coordinate $(i,j)$ of image $I$, $o_{i,j}^k$ is the result after neuron $k$, and finally $\vec{o}_{i,j}$ is the vector produced as output for this linear transformation. As studied in Linear Algebra, such transformation might map the domain into a contracted co-domain. Consecutive convolutional layers produce linear hyperplanes on their respective input spaces. Recall those following inputs are resultant of hyperplanes produced by previous layers.

Activation functions and/or pooling (sub-sampling) layers may be applied before convolutions. It is usual to employ an activation function to normalize the correlation results produced by a convolutional layer, helping to bound values (ensuring they do not tend to plus or minus infinity along iterations) and consequently in the convergence of the loss function. In this point, it is worth to mention the Multilayer Perceptron algorithm, which has its convergence improved when input data are normalized. That is due to values produced by function $f(\text{net})$ are maintained around a quasi-convex region, supporting the gradient descent algorithm~\cite{haykin1994neural}.

The Rectified Linear Unit (ReLU) is currently the most employed activation function in the context of CNNs.
ReLU avoids negative values and does not have an upper bound, thus allowing to represent magnitudes of input data as follows:
\begin{align*}
f(x) = \max(0,x).
\end{align*}

In terms of convolution, this means only positive values have provided a relevant response for filters, so the other information may be discarded. On the point of view of the Statistical Learning Theory, it is the same as having a linear hyperplane shattering a given input space and only one of the half spaces produced by such a function is taken, significantly reducing the data space.

In order to reduce the data space dimensionality, pooling layers are often employed by selecting the most relevant information according to some criterion.
It also provides some robustness to data invariance and distortion~\cite{scherer2010evaluation}.
The two main sub-sampling strategies applied with CNNs are: i) max-pooling, which keeps the maximum value of each local receptive field of image $\vec{I}_{i,j}$; and ii) average-pooling, which computes the average of each local receptive field of image $\vec{I}_{i,j}$.

In fact, the max-pooling operation results in an infinite norm:
\begin{align*}
o_{i,j} = \max(\vec{I}_{i,j}),
\end{align*}
in which $o_{i,j}$ receives the maximum value of the column vector $\vec{I}_{i,j}$ corresponding to some neighborhood of $I$, located at coordinate $(i,j)$.

As the infinite norm, its usage is justified when the statistical mode of data values is useful. It is also employed when one needs to measure the distance between data distributions~\footnote{For example, we here recall the Glivenko-Cantelli Theorem refers to the maximum distance between cumulative distributions.}.
On the other hand, the average-pooling is applied when $L_2$ norm results in a better space.
As consequence, depending on the target classification task, one of those sub-sampling strategies is more indicated. As an alternative of pooling operators, one can also increase the stride of the convolutions in order to produce spatial dimensionality reduction~\cite{Ponti2017everything, Springenberg2015striving}.

In addition, it is common to use Batch Normalization~\cite{Ioffe2015batch}, yet another linear transformation that shifts and scales the whole batch while it is propagated forward on the network.

Recent CNN architectures are composed of blocks performing: Convolution, Activation, Pooling and Batch Normalization, not necessarily in this order. At each block, the reduced space is linearly divided by the convolutional layer. At the end of the network, at least one or more fully connected layer is used, which work as in a Multilayer Perceptron. Those are responsible for training linear hyperplanes and finally perform classification at the output layer~\cite{Goodfellow2016deep, lecun2015deep}.

\section{A General Analysis on Convolutional Neural Networks}\label{sec:general}

In order to analyze the learning convergence for general-purpose Convolutional Neural Networks, we now set a given architecture to assess its Shattering coefficient, and, finally, understand how complex its class of admissible functions is, a.k.a. algorithm bias.
Consider a CNN architecture with the following layers: i) convolutional layer having $a$ neurons with $b \times c$ as filter size; ii) activation function followed by a max-pooling layer that reduces image dimensions by a factor of $2$; iii) convolutional layer having $d$ neurons with $e \times f$ as filter size; and, lastly, a single fully-connected layer with $g$ neurons.
Let $n$ input $\alpha \times \beta$ images be provided by this architecture during the training stage. As main question, we have: are those $n$ training examples enough to ensure this architecture has learned some concept?

To answer such a question, we now detail how training occurs. First of all, images are organized as local vectors with the same dimensions as the convolutional filter at the first layer, i.e., $b \times c$.
From this, we can use Algorithm~\ref{alg:shattering-estimator} to estimate the shattering coefficient $f_{\text{conv} 1}(n)$ for each neuron at such a layer. As there are $a$ neurons in this first layer, the shattering coefficient is $(f_{\text{conv} 1}(n))^a$ until now. The activation function and the max-pooling operations will help to reduce data dimensionality so images will become smaller, in fact $\frac{\alpha}{2} \times \frac{\beta}{2}$.
Then, local receptive fields of such images will be built up with the same dimensionality as the filters at the next convolutional layer, i.e., $e \times f$.
Then, Algorithm~\ref{alg:shattering-estimator} is again used to estimate the Shattering coefficient $f_{\text{conv} 2}(n)$ for each neuron of this second convolutional layer. As there are $d$ neurons, the layer Shattering coefficient is given by $(f_{\text{conv} 2}(n))^d$. If more layers are present, the same process must be repeated until all convolutional layers are considered.

As next step, we combine the Shattering coefficients of all layers using a product, because if a hyperplane is built in the first layer and another is produced by the second layer, then it is the same as having a combination of both in form:
\begin{align*}
f(n) = (f_{\text{conv} 1}(n))^a \times (f_{\text{conv} 2}(n))^d,
\end{align*}
consequently, if $k$ convolutional layers are used, then:
\begin{align*}
f(n) = \prod_{i=1}^{k} (f_{\text{conv} i}(n))^{|\text{conv} i|},
\end{align*}
in which $|\text{conv} i|$ is the number of neurons at layer $i$. Observe all this analysis is performed while disconsidering the fully-connected layers which add up more complexity to the Shattering coefficients. As matter of fact, we are only interested in analyzing the convolutional layers due to they are characteristic in CNNs.

From the Statistical Learning Theory, we have that learning, according to the Empirical Risk Minimization Principle~\cite{vonLuxburg,Scholkopf2002learning,Vapnik2013nature}, is only ensured if:
\begin{align*}
\lim_{n \rightarrow \infty} \frac{\log f(n)^k}{n} \approx 0,
\end{align*}
meaning this CNN architecture will require a greater training sample with size $n$ as the Shattering coefficient becomes more complex, as indeed happens by adding up more neurons at a layer or more layers. Answering our question.

Another interesting point for analysis is: how a single convolutional layer would be enough, in terms of admissible functions, to represent the classifiers as deep network such as CNN? In order to answer this, consider five convolutional layers with $a, b, c, d, e$ neurons each and the following Shattering coefficients for individual neurons at those layers:
\begin{align*}
f_{\text{conv} 1}(n) = \lambda_1 n^2,\\
f_{\text{conv} 2}(n) = \lambda_2 n^2,\\
f_{\text{conv} 3}(n) = \lambda_3 n^2,\\
f_{\text{conv} 4}(n) = \lambda_4 n^2,\\
f_{\text{conv} 5}(n) = \lambda_5 n^2,
\end{align*}
thus, the Shattering coefficient for such CNN is:
\begin{align*}
f(n) &= \prod_{i=1}^{k} (f_{\text{conv} i}(n))^{|\text{conv} i|}\\
&= (\lambda_1 n^2)^a \times (\lambda_2 n^2)^b \times (\lambda_3 n^2)^c \times (\lambda_4 n^2)^d \times (\lambda_5 n^2)^e\\
&= \lambda_1^a \lambda_2^b \lambda_3^c \lambda_4^d \lambda_5^e (n^2)^(a+b+c+d+e),
\end{align*}
therefore, a single convolutional layer should have $a+b+c+d+e$ neurons to provide an approximate similar space of admissible function, i.e.:
\begin{align*}
f_\text{single}(n) &= (\lambda_1 n^2)^(a+b+c+d+e)\\
&= \lambda_1^a \lambda_1^b \lambda_1^c \lambda_1^d \lambda_1^e (n^2)^(a+b+c+d+e),
\end{align*}
or the same bias if $\lambda_j = \lambda_k \, \forall j,k$. Provided the space of admissible functions is defined by the Shattering coefficient~\cite{Scholkopf2002learning}.

\section{Analyzing Specific Deep Neural Networks}\label{sec:analyzing}

In order to provide a practical framework to analyze Deep Learning architectures, we decided to consider AlexNet and VGG16 in this paper.
AlexNet is an implementation of a particular CNN architecture developed by Krizhevsky et al.~\cite{krizhevsky2012imagenet} in order to classify the ImageNet dataset. This architecture was empirically selected to extract features considering the object recognition task. According to the authors, a normalization layer was included to improve the network generalization, and also, an overlapping was applied on max-pooling layers to reduce the classification error rate, as well as, the overfitting.
The execution of CNN training was divided in two GPUs, in which each one is responsible to a half of convolutional layer neurons and fully-connected layers neurons, running simultaneously. It addition, data augmentation and dropout was employed to reduce the overfitting caused by this deep architecture. 

VGG16 was developed by Simonyan and Zisserman~\cite{simonyan2014very} in order to improve the results provided by AlexNet for the ImageNet dataset. As main goal, they intended to study how depth impacts on the network classification, concluding that increasing depth may improve results.
The main characteristic of this architecture is the small filter size used in convolutional layers.
VGG16 applies non-overlapping max-pooling layers and no normalization layer. It is worth to mention that both networks employ ReLU as the activation function right after each convolutional layer.

\subsection{AlexNet}


All discussion in this section is based on the AlexNet implementation from Caffe deep learning framework~\cite{jia2014caffe}, which has the following convolutional layers (Conv for short): i) Conv 1 -- $96$ neurons using filter size $11 \times 11$; ii) Conv 2 -- $256$ neurons with $5 \times 5$; iii) Conv 3 -- $384$ neurons with $3 \times 3$; iv) Conv 4 -- $384$ neurons with $3 \times 3$; and v) Conv 5 -- $256$ neurons with $3 \times 3$.
By using Algorithm~\ref{alg:shattering-estimator}, we estimated the Shattering coefficients for each single neuron at every layer, as follows:
\begin{align*}
\begin{split}
f_{\text{conv} 1}(n) &= 4806.3 n^2 -120396 n +758504,\\
f_{\text{conv} 2}(n) &= 2706.77 n^2 -66791.6 n + 417022,\\
f_{\text{conv} 3,4,5}(n) &=1077.13 n^2 -27910 n + 184905,
\end{split}
\end{align*}
whose squared of residuals and error percentages were, respectively:
\begin{align*}
\begin{split}
2609.20 & 8.885\%\\
2568.89 & 15.53\%\\
3902.56 & 15.77\%,
\end{split}
\end{align*}
as consequence, the overall Shattering coefficient for AlexNet is:
\begin{align*}
\begin{split}
f(n) &= (4806.3 n^2 -120396 n +758504)^{96} \times (2706.77 n^2 -66791.6 n + 417022)^{256} \times \\
&\quad (1077.13 n^2 -27910 n + 184905)^{384} \times (1077.13 n^2 -27910 n + 184905)^{384} \times \\
&\quad (1077.13 n^2 -27910 n + 184905)^{256},
\end{split}
\end{align*}
and the learning convergence only occurs when:
\begin{align*}
\begin{split}
\lim_{n \rightarrow \infty} & \frac{\log\{(4806.3 n^2 -120396 n +758504)^{96} \times (2706.77 n^2 -66791.6 n + 417022)^{256} \times}{} \\
&\quad  \frac{(1077.13 n^2 -27910 n + 184905)^{384} \times (1077.13 n^2 -27910 n + 184905)^{384} \times}{n} \\
&\quad \frac{(1077.13 n^2 -27910 n + 184905)^{256}\}}{} \approx 0,
\end{split}
\end{align*}
which can be simplified in terms of two functions forming a lower and upper bound for such Shattering, in form:

\begin{align*}
\begin{split}
  0 &< \lambda_1 n^{2752} \le f(n) \le \lambda_2 n^{2752}\\
  0 &< \lambda_1 n^{2752} \le (4806.3 n^2 -120396 n +758504)^{96} \times (2706.77 n^2 -66791.6 n + 417022)^{256} \times \\ 
  & \quad (1077.13 n^2 -27910 n + 184905)^{384} \times (1077.13 n^2 -27910 n + 184905)^{384} \times \\ 
  & \quad (1077.13 n^2 -27910 n + 184905)^{256} \le \lambda_2 n^{2752}\\
  0 &< \lambda_1 n^{2752} \le [ (4806.3 n^2)^{96} + \ldots + 758504^{96} ] \times [(2706.77 n^2)^{256} + \ldots + \\ 
  & \quad 417022^{256} ] \times [ (1077.13 n^2)^{384} + \ldots + 184905^{384} ] \times [ (1077.13 n^2)^{384} + \ldots + 184905^{384} ] \times \\ 
  & \quad [ (1077.13 n^2)^{256} + \ldots + 184905^{256} ] \le \lambda_2 n^{2752},
\end{split}
\end{align*}
which was opened using binomials. Then, dividing all terms by $n^{2752}$, we found:
\begin{align*}
\begin{split}
  0 < \lambda_1 & \le [ \frac{(4806.3 n^2)^{96}}{n^{2560}} + \ldots + \frac{758504^{96}}{n^{2752}} ] \times [ \frac{2706.77 n^2)^{256}}{n^{2752}} + \ldots + \frac{417022^256}{n^2752} ] \times \\ 
  & \quad [ \frac{(1077.13 n^2)^{384}}{n^2752} + \ldots + \frac{184905^{384}}{n^{2752}} ] \times [ \frac{(1077.13 n^2)^{384}}{n^{2752}} + \ldots + \frac{184905^{384}}{n^{2752}} ] \times \\ 
  & \quad [ \frac{(1077.13 n^2)^{256}}{n^{2752}} + \ldots + \frac{184905^{256}}{n^{2752}} ] \le \lambda_2,
\end{split}
\end{align*}
which clearly converges to some constant value so:
\begin{align*}
  \lambda_1 n^{2752} \le f(n) \le \lambda_2 n^{2752},
\end{align*}
for $0 < \lambda_1 \le \lambda_2$. Therefore, there are two functions with different values for $\lambda$ which envelope the Shattering, so that we can assume:
\begin{align*}
  \lim_{n \rightarrow \infty} \frac{\log\{ \lambda_1 n^{2752} \}}{n} \approx 0,
\end{align*}
for the best case scenario, i.e., for the lower limit of such Shattering. Learning occurs after the following convergence:
\begin{align*}
  \frac{\log\{ \lambda_1 n^{2752} \}}{n} \leq \gamma,
\end{align*}
which can be written as:
\begin{align*}
  \frac{\log{\lambda_1} + 2752 \log{n}}{n} \leq \gamma
\end{align*}
knowing $n \in \mathbb{N}$ and $\lambda_1 > 0$. Assuming $\gamma=0.01$ and $\lambda_1=10^{-323}$ (the smallest value the R Statistical Software can compute, so that we have the smallest possible influence provided by such constant), we would need a training sample with $n=4,117,104$ examples. In the same context, if we accept a lower learning guarantee such as $\gamma=0.1$, we would still need $n=343,347$ examples in the training set. Here comes the problem with deep networks, they require an expressively large training set. A question that comes up is: is AlexNet useful if trained and made available after analyzing a small training set? No.

Putting differently, consider the dataset ImageNet with $14,197,122$ images, and that all images would be provided to this architecture we have just analyzed.
What would be $\gamma$ for such training? In fact, we would have:
\begin{align*}
  \frac{\log{\lambda_1} + 2752 \log{n}}{n} \approx 3.19 \times 10^{-3},
\end{align*}
what is indeed enough to ensure learning. Now we analyze its Generalization Bound (Equation~\ref{eq:generalizationbound}) in form:
\begin{align*}
\begin{split}
    R(f_{\text{w}}) &\leq \underbrace{R_\text{emp} (f_{\text{w}})}_\text{training error} + \underbrace{\sqrt{-\frac{4}{n} \left( \log(\delta) - \log(2\mathcal{N}(\mathcal{F},2n)) \right)}}_{\text{divergence factor}\; \epsilon}\\
    R(f_{\text{w}}) &\leq \underbrace{R_\text{emp} (f_{\text{w}})}_\text{training error} + \underbrace{\sqrt{-\frac{4}{14,197,122} \left( \log(\delta) - \log(2 \times 3.19 \times 10^{-3}) \right)}}_{\text{divergence factor}\; \epsilon},
\end{split}
\end{align*}
and now assume $\delta = 0.05$, meaning we wish to ensure the empirical risk is a good estimator for the expected risk with probability equals to $95\%$ (as typically considered in Statistics):
\begin{align*}
    R(f_{\text{w}}) &\leq \underbrace{R_\text{emp} (f_{\text{w}})}_\text{training error} + \underbrace{\sqrt{-\frac{4}{14,197,122} \left( \log(0.05) - \log(2 \times 3.19 \times 10^{-3}) \right)}}_{\text{divergence factor}\; \epsilon}\\
    R(f_{\text{w}}) &\leq \underbrace{R_\text{emp} (f_{\text{w}})}_\text{training error} + \underbrace{\sqrt{-\frac{4}{14,197,122} \left( \log(0.05) - \log(2 \times 3.19 \times 10^{-3}) \right)}}_{\text{divergence factor}\; \epsilon}\\
    R(f_{\text{w}}) &\leq \underbrace{R_\text{emp} (f_{\text{w}})}_\text{training error} + 0.0007616277 i,
\end{align*}
finally, confirming any measure we have obtained using a $k$-fold cross validation strategy is enough to conclude about the empirical risk, as well as it is significant to ensure learning, given the expected risk $R(f_{\text{w}})$ for the worst classifier $f_{\text{w}}$ is bounded. The imaginary number is due to the root of a negative value, meaning that this whole formulation is valid for a smaller value than $\delta = 0.05$. In fact, such probability can assume values $\delta < 2 \times 3.19 \times 10^{-3} = 0.00638$, what is the same as say that this formulation ensures the empirical risk is a good estimator for the expected one with probability $1-\delta = 0.99362$, i.e., for $99.36\%$ of cases.

Now consider someone has trained this same architecture using the MNIST dataset, which has $70,000$ images.
In that situation, the best $\gamma$ would be:
\begin{align*}
  \frac{\log{\lambda_1} + 2752 \log{n}}{n} \approx 0.4279751,
\end{align*}
even setting $\lambda_1=10^{-323}$ which is numerically computable. In that situation, the learning guarantee would be given by:
\begin{align*}
    R(f_{\text{w}}) &\leq \underbrace{R_\text{emp} (f_{\text{w}})}_\text{training error} + \underbrace{\sqrt{-\frac{4}{70,000} \left( \log(0.05) - \log(2 \times 0.4279751) \right)}}_{\text{divergence factor}\; \epsilon}\\
    R(f_{\text{w}}) &\leq \underbrace{R_\text{emp} (f_{\text{w}})}_\text{training error} + 0.01273957,
\end{align*}
and, therefore, different algorithms providing distinct empirical risks (one minus the accuracy), even while using $k$-fold cross validation, might be statistically similar given the error of $1.27\%$ associated to the Shattering coefficient.
Consequently, if two classification algorithms provide the same empirical risk, the one whose Shattering is less complex is the best.

In addition, consider two algorithms providing the following empirical risks:
\begin{align*}
    R_\text{emp} (f_{\text{w}})_\text{alg1} &= 0.01\\
    R_\text{emp} (f_{\text{w}})_\text{alg2} &= 0.015,
\end{align*}
and whose divergence factor is given by $0.01273957$ and $\frac{0.01273957}{2}$, respectively. Observe, in fact, that the expected risks for both will be:
\begin{align*}
    R(f_{\text{w}})_\text{alg1} &\leq 0.01 + 0.01273957 = 0.02273957\\
    R(f_{\text{w}})_\text{alg2} &\leq 0.015 + \frac{0.01273957}{2} = 0.02136978,
\end{align*}
what confirms the empirical risk typically analyzed in deep learning contests is not enough to prove which algorithm is indeed the best.

\subsection{VGG16}

The implementation of VGG16 training in Caffe deep learning framework~\cite{jia2014caffe} has the following convolutional layers (Conv for short):
i) Conv 1 and 2 -- $64$ neurons using filter size $3 \times 3$;
ii) Conv 3 and 4 -- $128$ neurons with $3 \times 3$;
iii) Conv 5, 6, 7 and 8 -- $256$ neurons with $3 \times 3$;
iv) Conv 9, 10, 11, 12, 13 and 14 -- $512$ neurons with $3 \times 3$.
Also using Algorithm~\ref{alg:shattering-estimator}, the Shattering coefficients estimated for single neurons at each layer are the same due to the convolutional filter size is always $3 \times 3$:
\begin{align*}
    g_{\text{conv}}(n) = 1077.13 n^2 -27910 n + 184905,
%
%
\end{align*}
whose squared of residual and error percentage were $3902.56$ and $15.77\%$, respectively.
As consequence, the overall Shattering coefficient for VGG16 is:
\begin{align*}
\begin{split}
    g(n) &= (1077.13 n^2 -27910 n + 184905)^{64} \times (1077.13 n^2 -27910 n + 184905)^{64} \times \\ 
  & \quad (1077.13 n^2 -27910 n + 184905)^{128} \times (1077.13 n^2 -27910 n + 184905)^{128} \times \\ 
  & \quad (1077.13 n^2 -27910 n + 184905)^{256} \times (1077.13 n^2 -27910 n + 184905)^{256} \times \\ 
  & \quad (1077.13 n^2 -27910 n + 184905)^{256} \times (1077.13 n^2 -27910 n + 184905)^{256} \times \\ 
  & \quad (1077.13 n^2 -27910 n + 184905)^{512} \times (1077.13 n^2 -27910 n + 184905)^{512} \times \\ 
  & \quad (1077.13 n^2 -27910 n + 184905)^{512} \times (1077.13 n^2 -27910 n + 184905)^{512} \times \\ 
  & \quad (1077.13 n^2 -27910 n + 184905)^{512} \times (1077.13 n^2 -27910 n + 184905)^{512},
\end{split}
\end{align*}
and the learning convergence only occurs when:
\begin{align*}
  \lim_{n \rightarrow \infty} \frac{\log\{(1077.13 n^2 -27910 n + 184905)^{4480}\}}{n} \approx 0,
\end{align*}
which opened using binomials provides the following:
\begin{align*}
  \lim_{n \rightarrow \infty} \frac{\log\{1077.13 n^{8960} + \ldots + 184905^{4480}\}}{n} \approx 0,
\end{align*}
which can also be simplified in terms of two lower and upper bound functions:
\begin{align*}
  0 &< \theta_1 n^{8960} \le g(n) \le \theta_2 n^{8960}\\
  0 &< \theta_1 n^{8960} \le (1077.13 n^2)^{4480} + \ldots + 184905^{4480} \le \theta_2 n^{8960},\\
  \text{}
\end{align*}
in binomial-open form. So, dividing all terms by $n^{8960}$ we have:
\begin{align*}
  0 < \theta_1 \le [ \frac{ (1077.13 n^2)^{4480} }{n^{4480}} + \ldots + \frac{184905^{4480}}{n^{4480}} ] \le \theta_2,
\end{align*}
which clearly converges to some constant value, so that:
\begin{align*}
  \theta_1 n^{8960} \le g(n) \le \theta_2 n^{8960},
\end{align*}
for $0 < \theta_1 \le \theta_2$. As consequence, there are two functions with different values for $\theta$ which envelope the Shattering, so we can assume:
\begin{align*}
  \lim_{n \rightarrow \infty} \frac{\log\{ \theta_1 n^{8960} \}}{n} \approx 0,
\end{align*}
for the best case scenario, i.e., for the lower limit of such Shattering (lowest complexity).
As performed while analyzing AlexNet, learning only occurs after the following convergence:
\begin{align*}
  \frac{\log{\theta_1} + 8960 \log{n}}{n} \leq \gamma,
\end{align*}
knowing $n \in \mathbb{N}$ and $\theta_1 > 0$. As previously performed for AlexNet, assume $\gamma=0.01$ and $\theta_1=10^{-323}$~\footnote{The smallest value the R Statistical Software can compute, so that we have the smallest possible influence provided by such constant.}, we would need a training sample with $n=14,713,454$ examples. Even accepting a lower learning guarantee such as $\gamma=0.1$, $n=1,250,459$ examples are required to compose the training set.

Assuming ImageNet is used as input dataset, $\delta=0.05$ as previously discussed, and $\theta_1=10^{-323}$ which is still numerically computable using the R Statistical Software, we would have $\gamma = 0.01039353$ and the learning guarantee would be given by:
\begin{align*}
    R(f_{\text{w}}) &\leq \underbrace{R_\text{emp} (f_{\text{w}})}_\text{training error} + \underbrace{\sqrt{-\frac{4}{14,197,122} \left( \log(0.05) - \log(2 \times 0.01039353) \right)}}_{\text{divergence factor}\; \epsilon}\\
    R(f_{\text{w}}) &\leq \underbrace{R_\text{emp} (f_{\text{w}})}_\text{training error} + 0.000497280 i,
\end{align*}
and therefore learning is ensured for such dataset, specially due to its number of examples. This again confirms the error measured with the $k$-fold cross validation strategy is enough to conclude about the empirical risk, as well as it is significant to ensure learning, given the expected risk $R(f_{\text{w}})$ for the worst classifier $f_{\text{w}}$ is bounded. We again have an imaginary number as result, due to this formulation is valid for a smaller value than $\delta = 0.05$. Such probability can assume values $\delta < 2 \times 0.01039353 = 0.02078706$, what is the same as say that the empirical risk is a good estimator for the expected one with probability $1-\delta = 0.9792129$, i.e., for $97.92\%$ of cases.

\subsection{Discussion}\label{sec:discussion}

During our studies, we have confirmed most Shattering coefficient for single neurons are satisfactorily estimated using a second-order polynomial, while the number of neurons per layer and along layers is what mostly influences in the algorithm bias. While analyzing all neurons even using different convolutional filter sizes, we observe their shatterings are still in form $a n^2$, while the total number of neurons per layer makes such complexity to become:
\begin{align*}
    (a n^2)^{\alpha},
\end{align*}
and by adding $l$ layers, we have:
\begin{align*}
    (a n^2)^{\alpha_1} \times (a n^2)^{\alpha_2} \times \ldots \times (a n^2)^{\alpha_l},
\end{align*}
highly impacting on the number of training examples necessary to ensure learning according to the Generalization Bound (Equation~\ref{eq:generalizationbound}).

For example, Simonyan and Zisserman~\cite{simonyan2014very} mention the VGG16 kernel masks were reduced in attempt to make the algorithm bias smaller too, improving learning. However, after the results showed in this paper, we conclude the number of neurons still causes greater influences in the overall bias complexity of those classification algorithms.

The same authors state a three-layer CNN with $64$ neurons at each layer and mask sizes $3 \times 3$ provides the same functions as a single-layer CNN with $64$ neurons but using a convolutional filter size equals to $7 \times 7$. They also mention that is easy to be proved, but no proof is provided.
Following our theoretical framework, we found the following Shattering coefficient for single neurons using $3 \times 3$ and $7 \times 7$:
\begin{align*}
    f_{3 \times 3}(n) = 7.85 n^2 + 147.69 n - 2871.02 \\
    f_{7 \times 7}(n) = 9.81 n^2 + 11.32 n - 171.91
\end{align*}
From that, the $3 \times 3$ three-layer CNN will have the following Shattering coefficient:
\begin{align*}
    f(n) & = ((7.85 n^2 + 147.69 n - 2871.02)^{64})^3\\
    f(n) & = ((7.85 n^2 + 147.69 n - 2871.02)^{192}\\
         & = 10^{-384} (785 n^2 + 14769 n - 287102)^{192}
\end{align*}
while the $7 \times 7$ single-layer has:
\begin{align*}
    g(n) & = (9.81 n^2 + 11.32 n - 171.91)^{64}\\
         & = 10^{-128} (981 n^2 + 1132 n - 17191)^{64}
\end{align*}
what turns out to be very different in terms of Shattering coefficients, which is a confident measurement for the complexity of the admissible class of functions. We recall such measurement is theoretically employed to understand the number of distinct classifiers a given algorithm is capable of inferring~\cite{vonLuxburg,Vapnik2013nature,Scholkopf2002learning}. Therefore, both deep learning architecture are indeed distinct from each other.

From that, Simonyan and Zisserman~\cite{simonyan2014very} may concluded both architectures provide similar results simply based on the empirical risks observed in practical applications. However, their biases are very different. Such a conclusion is probably due to even the less complex bias, measured through the Shattering coefficient $g(n)$, was already enough to learn the input data. Consequently, the other architecture was unnecessarily complex.

\section{Conclusions}\label{sec:conclusions}

This paper proposed an algorithm to estimate the Shattering coefficient of neurons that compose the layers of Convolutional Neural Networks, therefore measuring the complexity of the algorithm bias, i.e., the size of its space of admissible functions. Then, we used it to study the learning convergence of Convolutional Neural Networks. This was motivated due to the criticism some researchers have against DL as well as the indiscriminate usage of such architectures to find solutions for any classification task. In fact, we believe DL is very important however instead of simply analyzing accuracy/error results, we discussed in this paper about the importance in understanding its theoretical learning bounds.

The Statistical Learning Theory (SLT) was central to the development of this article, particularly the Empirical Risk Minimization Principle which relies on the Law of Large Numbers to define learning guarantees in the context of supervised learning. The conclusions drawn in this paper are strongly based on Convolutional Neural Networks, specially because they have been used as a standard for deep learning applications. We started proving CNNs built up linear transformations at each layer as well as every single neuron is responsible for a linear hyperplane dividing the data space into two regions. Next, we show how single versus multiple layers influence in the Shattering coefficient for a general-purpose architecture. Later, we studied two well-known architectures AlexNet and VGG16. We confirmed that by having enough training examples, we can indeed prove learning convergence as, thus, employ CNNs in practice. On the other hand, if training examples are not enough, learning cannot be ensured, and results may be found by chance. In addition, we show how the Shattering coefficient makes the simple study of empirical risks insufficient to take conclusions on which is the best CNN architecture to tackle a given problem. At last, we compared different topologies to conclude whether they provide the same space of admissible functions, and it is clear they do not, besides some authors make such a claim.

\bibliography{dnn_refs}
\bibliographystyle{plain}

\end{document}